\documentclass[letterpaper]{article} 
\usepackage{aaai2027}  
\usepackage[hyphens]{url}  
\usepackage{graphicx} 
\usepackage{natbib}  
\usepackage{caption} 
\usepackage{algorithm}
\usepackage{algorithmic}

\usepackage{newfloat}
\usepackage{listings}
\DeclareCaptionStyle{ruled}{labelfont=normalfont,labelsep=colon,strut=off} 
\floatstyle{ruled}
\newfloat{listing}{tb}{lst}{}
\floatname{listing}{Listing}

\usepackage{booktabs}
\usepackage{multirow}
\usepackage{array}      
\usepackage{colortbl}   
\usepackage{multirow}   
\usepackage{booktabs}   
\usepackage{graphicx}   
\usepackage{xcolor}     
\usepackage{amsmath}

\usepackage{amssymb}   
\usepackage{colortbl}  

\usepackage[utf8]{inputenc}
\usepackage{booktabs}   
\usepackage{colortbl}   
\usepackage{xcolor}     
\usepackage{array}      
\usepackage{graphicx}   
\usepackage{colortbl}
\usepackage{xcolor}
\usepackage{multirow}
\usepackage{svg}
\usepackage{algorithm}
\usepackage{algorithmic}

\title{GeoForge: Non-Parametric Self-Evolving Agents for Earth-Observation Reasoning}

\author{
    Xin Xiao\textsuperscript{1},
    Jiang Zhong\textsuperscript{1$\dagger$},
    Junnan Zhu\textsuperscript{2} \\
    Yingchao Feng\textsuperscript{3},
    Peijin Wang\textsuperscript{3},
    Yidan Zhang\textsuperscript{3},
    Kaiwen Wei\textsuperscript{1$\dagger$}
}
\affiliations{
    \textsuperscript{\rm 1}School of Computer Science, Chongqing University \\
    \textsuperscript{\rm 2}MAIS, Institute of Automation, Chinese Academy of Sciences \\
    \textsuperscript{\rm 3}Aerospace Information Research Institute, Chinese Academy of Sciences
}

\begin{document}

\maketitle

\begin{abstract}
Earth observation (EO) agents construct scientifically valid tool workflows and ground their conclusions in current geospatial evidence. This is challenging because EO workflows are constrained by sensing semantics, product dependencies, spatial and temporal compatibility, and parameter requirements. Existing agents often search a broad operation space for each query, while recent self-evolving systems do not fully organize heterogeneous EO trajectories into reusable knowledge across different decision levels. To solve this problem, we present GeoForge, a training-free, self-evolving framework that transforms completed trajectories into a structured nonparametric execution state. GeoForge constrains the operation space according to the sensing context, then retrieves a task-conditioned prior from three complementary memories. Workflow Graph Memory captures global operation order, Action-Level Experiences provide local corrections, and the Adapted Skill Standard Operating Procedure preserves procedural and data constraints. The retrieved prior guides tool execution, while current observations remain the basis of the final answer. After each task, a safety-gated distillation process converts grounded trajectories into reusable execution knowledge for future retrieval. This execution, distillation, and reuse loop improves planning without updating the backbone LLM. Experiments on multiple geospatial benchmarks demonstrate that GeoForge consistently improves both task accuracy and tool-use trajectory quality across diverse LLM backbones, while substantially reducing tool-planning and reasoning errors for most LLMs.

\end{abstract}

\section{Introduction}

Large language model (LLM) agents can interpret user requests, plan complex intermediate actions, and interact with external tools to solve complex tasks \citep{yao2022react,wu2023autogen,shinn2024reflexion}. This capability is particularly promising for Earth Observation (EO), where scientific analysis requires coordinated reasoning over multimodal observations, geospatial products, and specialized tools \citep{wang2024gpt,xiao2025foundation}. EO agents can automate diverse real-world analytical pipelines by selecting observations, generating products, and grounding conclusions with geospatial evidence, thereby accelerating scientific discovery \citep{van2025opportunities,wang2026large}.

\begin{figure}[t]
\centering
\includegraphics[width=0.9\columnwidth]{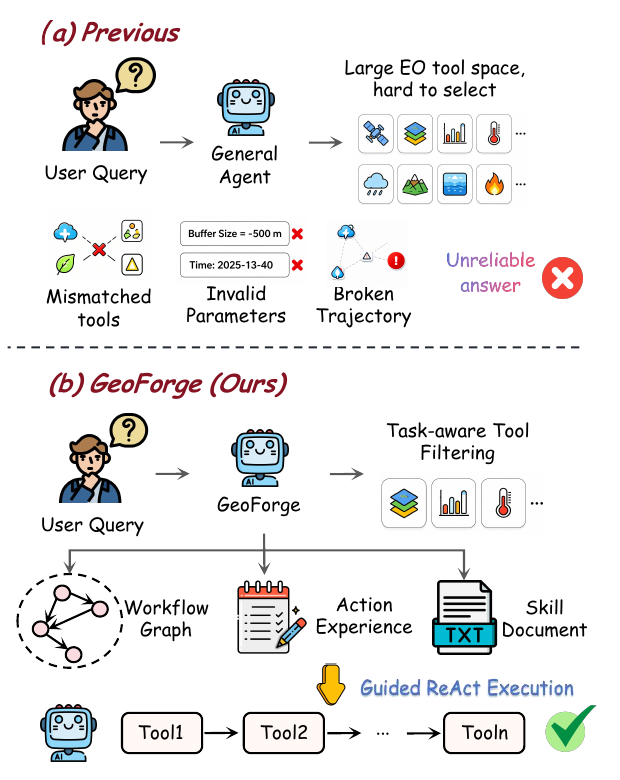} 
\caption{
Comparison of traditional EO agents and GeoForge. (a) Traditional agents fail with mismatched tools and broken trajectories. (b) GeoForge uses self-evolving and task-aware filtering for reliable EO task execution.
}
\label{fig:introduction}
\end{figure}

Reliable EO analysis is governed by sensing semantics, observation conditions, and workflow dependencies \citep{xiao2025foundation,wang2024gpt}. Unlike general tool-use scenarios, EO workflows depend on available observations, sensing modalities, and intermediate products rather than task descriptions alone \citep{zhu2017deep}. Without such domain constraints, agents may select irrelevant operations, misuse inputs or products, and construct invalid workflows. Since these rules vary across tasks and sensing contexts, they are difficult to encode in static prompts, motivating task-conditioned execution knowledge that captures scientific compatibility and can be reused across tasks while continuously adapting to newly encountered sensing scenarios.

Existing EO agents commonly adopt general reasoning and action frameworks, in which an LLM constructs workflows through iterative reasoning and tool interaction \citep{xu2024rs,feng2025earth}. Recent benchmarks have demonstrated the potential of this paradigm, but also revealed persistent limitations in tool selection, execution order, argument grounding, and workflow completion \citep{shabbir2025thinkgeo,li2025designing}. As illustrated in Figure~\ref{fig:introduction}, exposing the agent to a large heterogeneous EO tool space without explicit execution guidance may lead to mismatched operations, invalid parameters, and broken trajectories. Without reusable priors about scientific compatibility and valid operation order, the agent must explore a broad action space for each query, resulting in redundant calls, incompatible operations, and incomplete workflows.

Related self-evolving geospatial systems have explored how external knowledge and interaction experience can improve geospatial problem solving without updating the backbone LLM. GeoEvolve~\citep{luo2025geoevolve} uses knowledge-guided code evolution to discover and refine task-specific geospatial algorithms, while GeoEvolver distills successful interactions and failure attributions into an evolving experience bank for fine-grained EO tool execution \citep{dai2026experience}. These studies demonstrate two complementary routes to self-evolution. GeoEvolve evolves algorithm implementations, whereas GeoEvolver accumulates interaction-level tool expertise. A distinct challenge remains in organizing reusable execution knowledge across an EO workflow as a whole, from global operation order to local action corrections and procedural constraints. Raw trajectories preserve excessive instance-specific detail, while compact textual summaries may omit the product dependencies, parameter conditions, and execution structure that determine workflow validity. The key challenge is therefore to transform completed trajectories into execution priors that are domain-valid, structured, task-conditioned, and safely updatable.

To address this challenge, we propose GeoForge, a training-free, self-evolving framework that maintains a structured nonparametric execution state outside the backbone LLM. Given a task, GeoForge first infers the sensing context to constrain the operation space and provide execution guidance. It then constructs a task-conditioned execution prior from three complementary memories: Workflow Graph Memory captures global workflow structures, Action-Level Experiences provide local corrections, and the Adapted Skill SOP preserves reusable procedures, data requirements, and parameter constraints. Together, these components guide operation selection and composition while grounding final answers in current observations and geospatial products. After each task, GeoForge distills grounded trajectories into reusable workflow structures, action-level lessons, and procedural knowledge, enabling continual improvement without updating model parameters. Experiments on multiple geospatial benchmarks demonstrate that GeoForge substantially improves task accuracy and tool-use trajectory quality over existing EO agents, while reducing tool-planning and reasoning errors across most backbone LLMs. The contributions are summarized as follows:

\begin{itemize}

\item We propose GeoForge, a training-free, self-evolving Earth observation agent framework. By continuously distilling remote-sensing execution trajectories into structured knowledge representations, GeoForge enables sustained planning and tool-use advancement across heterogeneous geospatial modalities without perturbing backbone LLM weights.

\item We introduce a three-level non-parametric memory together with a task-aware execution strategy, combining modality-aware tool filtering, Workflow Graph Memory, Action-Level Experiences, and an Adapted Skill SOP to provide structured planning priors and support continual self-evolution.

\item Extensive experiments on multiple Earth observation agent benchmarks demonstrate that GeoForge consistently improves both task accuracy and execution trajectory quality across diverse LLM backbones, validating the effectiveness for reliable EO tool use.

\end{itemize}

\section{Related Work}
\subsection{Agents for Earth Observation}

Recent advances in large language models have catalyzed tool-augmented agents for geospatial and Earth observation applications. Early efforts like Remote Sensing ChatGPT \cite{guo2024remote} and RS-Agent \cite{xu2024rs} demonstrated chaining perception and analysis modules under LLM planners, yet they reported final accuracy without structured evaluation of tool-use trajectories. Subsequent benchmarks raised evaluation rigor: ThinkGeo \cite{shabbir2025thinkgeo} introduced tasks requiring detection, segmentation, and geospatial reasoning; OpenEarthAgent \cite{shabbir2026openearthagent} extended this with a broader tool ecosystem. TerraBench \cite{nguyen2026terrabench} unified EO imagery, gridded environmental data, GIS reasoning, and simulation into a single framework with process-level metrics, revealing argument grounding and numeric reasoning as bottlenecks. Methodological innovations include Geo-OLM \cite{stamoulis2025geo} with state-driven reasoning that decouples task progression from tool calling, and Earth-Agent \cite{feng2025earth} with a ReAct-style framework integrating specialized tools. Hierarchical task abstraction \cite{li2025designing} has also been explored. TerraLogic \cite{yan2026terralogic} introduced a benchmark for hierarchical reasoning with error-aware replanning across optical, SAR, and IR modalities. 

\subsection{Self-Evolving Agents}

A parallel line of research explores agent improvement from interaction experience without parameter updates. ReAct \cite{yao2022react} demonstrated iterative reasoning-acting loops, with surveys organizing this field \cite{tao2024survey}. Memento introduced Read-Write Reflective Learning, where episodic memory stores outcomes and retrieval enables policy improvement \cite{zhou2025memento}; Memento-Skills treats reusable skills as artefacts that evolve through reflective revision \cite{zhou2026memento}. MemSkill reframes memory extraction as learnable skills with a controller-selector-designer architecture \cite{zhang2026memskill}. Trajectory-Informed Memory Generation extracts actionable learnings from execution trajectories \cite{fang2026trajectory}. SAGE introduces a self-evolving graph-memory engine \cite{wang2026sage}, and XSkill proposes continual learning from experience in multimodal agents \cite{jiang2026xskill}. These approaches show that continual learning need not reside in model weights; a self-improving memory or skill library can serve as persistent intelligence. 
Recent efforts have begun extending these ideas to Earth observation, including GeoEvolve, which automates geospatial model discovery through multi-agent collaboration \cite{luo2025geoevolve}, and Experience-Driven Multi-Agent Systems\cite{dai2026experience}.
However, they do not explicitly optimize tool-space exploration and workflow reuse for heterogeneous EO reasoning. GeoForge addresses this gap through task-aware tool constraints, retrieval-augmented planning, and continual non-parametric memory updates.


\section{Methodology}

\subsection{Overview}
\begin{figure*}[t]
\centering
\includegraphics[width=1.0\textwidth]{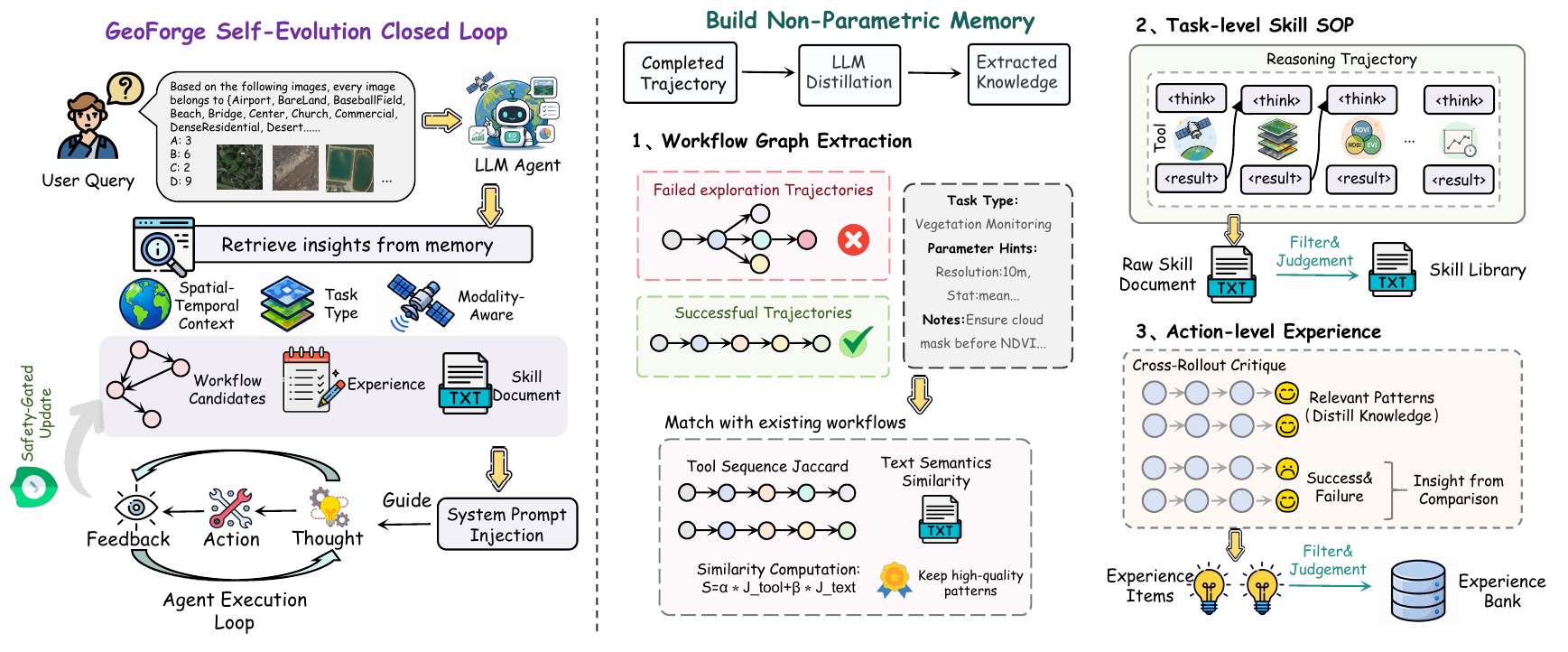} 
\caption{
Pipeline of GeoForge. The framework comprises three modules: a ReAct-based self-evolution closed loop, non-parametric memory construction with trajectory filtering and hybrid similarity matching, and multi-level knowledge abstraction for task skills and action experiences.
}
\label{fig1}
\end{figure*}

As shown in Figure~\ref{fig1}, we introduce GeoForge, a self-evolving EO agent that learns procedural knowledge for remote-sensing analysis without updating the weight of backbone. Workflow Graph Memory, Action-Level Experience, and an Adapted Skill SOP jointly guide EO data inspection, geospatial product construction, spatiotemporal reasoning, and argument grounding. After each episode, safety-gated trajectory distillation updates this external state.

\subsection{Problem Formulation}

An EO task is $x=(q,d,\mathcal{Y})$, comprising a scientific request, data collection, and answer space. Given an operation library $\mathcal{T}=\{T_i\}_{i=1}^{n}$, the agent selects either $a_t=(T_i,\theta_t)$ with $\theta_t\in\Omega_i$, or a prediction $y\in\mathcal{Y}$. A trajectory is
\begin{equation}
    \tau=(q,a_1,o_1,\ldots,a_m,o_m,y),
    \quad o_t=T_i(\theta_t),
    \label{eq:trajectory}
\end{equation}
Observations include EO data inventories, derived geospatial products, perception outputs, regional or temporal statistics, and execution errors. A valid trajectory must respect remote-sensing dataflow: relevant observations and modalities are identified before product generation, returned artifacts are propagated to downstream tools, and aggregation is performed only after checking spatial and temporal compatibility. Final answers must be supported by the current EO data rather than retrieved memory.

Let $z=[T(a_1),\ldots,T(a_m)]$ and $\Theta_\tau=[\theta_1,\ldots,\theta_m]$ denote the operation and argument traces. We evaluate final correctness, operation coverage and order, argument grounding, repetition, and incompatible transitions. Without expert trajectories at inference time, GeoForge approximates these execution priors through its external state.

\subsection{Workflow Graph Memory}

We represent high-level execution knowledge as a directed, reliability-aware graph $\mathcal{G}=(\mathcal{W},\mathcal{A},\mathcal{U})$ of workflow nodes, tool-transition edges, and usage statistics. Unlike a generic API graph, it encodes EO dependencies such as data discovery $\rightarrow$ modality-aware input organization $\rightarrow$ geospatial product generation $\rightarrow$ spatial or temporal aggregation $\rightarrow$ scientific interpretation. A node is
\begin{equation}
    w_i=(c_i,r_i,\mathbf{z}_i,\mathbf{p}_i,\mathbf{b}_i,Q_i,n_i^+,n_i^-),
    \label{eq:workflow_node}
\end{equation}
where $c_i$ is the task type, $r_i$ the workflow description, $\mathbf{z}_i$ an order-preserving compressed tool sequence, $\mathbf{p}_i$ parameter hints, $\mathbf{b}_i$ cautions, $Q_i$ provenance, and $n_i^+,n_i^-$ success/failure counts. Removing repeated tool names separates reusable workflow structure from data-expanded repetition.

For retrieval, each workflow is converted into a heterogeneous text-symbol object containing task type, summary, tool sequence, parameter hints, and cautions. Given a query $q$, the graph module infers a coarse EO task type $\hat{c}(q)$ and scores workflow nodes by combining two complementary similarity measures: \textit{tool sequence Jaccard} and \textit{text semantics similarity}. The overall score is computed as
\begin{equation}
    \begin{aligned}
        s_G(q,w_i)=
        &\,\alpha \cdot J_{\text{tool}}(q,w_i)
        +\beta \cdot J_{\text{text}}(q,w_i) \\
        &+\lambda_C\mathbf{1}[c_i=\hat{c}(q)]
        -\lambda_L\max(0,|\mathbf{z}_i|-L_0),
    \end{aligned}
    \label{eq:workflow_score}
\end{equation}
where 
\begin{equation}
    J_{\text{tool}}(q,w_i)=\frac{|V_{\text{tool}}(q)\cap V_{\text{tool}}(w_i)|}{\max(|V_{\text{tool}}(q)|,1)}
\end{equation}
measures the overlap of tool sequences, and
\begin{equation}
    J_{\text{text}}(q,w_i)=\frac{|V_{\text{text}}(q)\cap V_{\text{text}}(w_i)|}{\max(|V_{\text{text}}(q)|,1)}
\end{equation}
captures lexical affinity between the query and workflow description. The weighting coefficients $\alpha$ and $\beta$ balance the contribution of structural and semantic similarities, with $\alpha+\beta=1$. The term $\lambda_C$ rewards task-type consistency, while $-\lambda_L$ penalizes overly long tool sequences to favor parsimonious workflows.

Only nodes above score and support thresholds are serialized with their tool order, summary, parameter hints, and cautions. They are marked as planning priors, preventing past trajectories from replacing current observations.

\subsection{Action-Level Experiences}

The experience bank stores fine-grained execution knowledge as non-parametric corrections for local decisions under previously observed conditions. An item $e_i\in\mathcal{E}$ is
\begin{equation}
    e_i=(\chi_i,\alpha_i,\mathbf{a}_i,\rho_i,m_i),
    \label{eq:experience_def}
\end{equation}
where $\chi_i$ is a trigger, $\alpha_i$ a correction, $\mathbf{a}_i$ the source operations, $\rho_i$ provenance, and $m_i$ metadata. The bounded pair $(\chi_i,\alpha_i)$ corrects local EO decisions such as modality and time alignment, parameter grounding, reuse of returned geospatial artifacts, handling of spatial incompatibility, or stopping after sufficient evidence is obtained.

Given a query $q$, retrieval is formulated as a gated relevance model. Let $V(\cdot)$ be a domain-aware lexical projection and $\hat{c}(\cdot)$ a coarse task abstraction. We score each experience by:
\begin{equation}
    s_E(q,e_i)=
    \underbrace{\frac{|V(q)\cap V(\chi_i\Vert\alpha_i\Vert\rho_i)|}{\max(|V(q)|,1)}}_{\text{lexical affinity}}
    +\lambda_E
    \underbrace{\mathbf{1}[\hat{c}(\chi_i\Vert\rho_i)=\hat{c}(q)]}_{\text{task consistency}}.
    \label{eq:experience_score}
\end{equation}

The retrieved set is obtained under an evidence constraint:
\begin{equation}
\begin{split}
    \mathcal{E}_q=
    \operatorname{TopK}_{e_i\in\mathcal{E}}
    \Bigl[ & s_E(q,e_i)\cdot
    \mathbf{1}\big(|V(q)\cap V(e_i)|>0\big) \\
    & \cdot \mathbf{1}\big(|\mathbf{a}_i|>0\big) \Bigr].
\end{split}
\label{eq:experience_retrieval}
\end{equation}

The last indicator excludes free-form reflections not grounded in executable traces. The rewrite-and-filter operator $R_{\Theta}$ contextualizes retrieved experiences against the task and adapted SOP, preserving their condition--action form while removing inapplicable assumptions.

The bank evolves through novelty-preserving consolidation. Given candidates $\Delta\mathcal{E}$, GeoForge distills \textit{relevant patterns} from successful trajectories as condition--action rules, and derives \textit{insights from comparison} by contrasting successes and failures to attribute errors (e.g., modality mismatch, incorrect parameters, premature stopping) to corrective conditions. It then normalizes each item with $\nu(\cdot)$ and admits only entries outside existing equivalence classes:
\begin{equation}
    \mathcal{E}_{t+1}=
    \operatorname{Tail}_{N_E}
    \left(
    \mathcal{E}_{t}
    \cup
    \{e\in\Delta\mathcal{E}: \nu(e)\notin\nu(\mathcal{E}_t)\}
    \right).
    \label{eq:experience_update}
\end{equation}

This rule favors high-precision reusable corrections over idiosyncratic episodic details, ensuring that the experience bank accumulates generalizable knowledge from both positive and negative execution outcomes.

\subsection{Adapted Skill SOP}

Whereas experiences provide local corrections, the SOP specifies reusable task decompositions, intermediate artifacts, and evidence aggregation. We write
\begin{equation}
    \mathcal{S}=\{s_j\}_{j=1}^{M},\qquad
    s_j=(\mu_j,\omega_j,\pi_j,\beta_j,\kappa_j),
    \label{eq:skill_def}
\end{equation}
where $\mu_j$, $\omega_j$, $\pi_j$, $\beta_j$, and $\kappa_j$ encode the EO task, processing workflow, data and argument constraints, sensor/modality semantics, and failure-prevention rules.

Directly injecting $\mathcal{S}$ would create a high-recall but low-precision prompt. GeoForge instead constructs a task-conditioned SOP through an adaptation operator
\begin{equation}
    \mathcal{S}_q=A_{\Theta}(\mathcal{S},q,\mathcal{E}_q;L_S),
    \label{eq:skill_adapt}
\end{equation}
here, $A_{\Theta}$ is an LLM compressor bounded by $L_S$. Constrained generation preserves relevant workflows and cautions while removing irrelevant and instance-specific details, yielding a compact task-local SOP.
The adapted SOP aligns the retrieved graph $\mathcal{G}_q$ with contextualized experiences $\widetilde{\mathcal{E}}_q$. The graph supplies order, the SOP procedural semantics, and experiences local corrections, while execution evidence remains authoritative.
The SOP is updated by replacement-style consolidation rather than append-only accumulation. Given a distilled candidate $\Delta\mathcal{S}$, GeoForge accepts it only if it is complete, compact, grounded in an executable trajectory, and passes the safety gate in Eq.~\ref{eq:update_gate}. The update is
\begin{equation}
    \mathcal{S}_{t+1}=
    \begin{cases}
    \operatorname{Sanitize}(\Delta\mathcal{S}), & \psi(\tau,\hat{y})=1,\\
    \mathcal{S}_{t}, & \psi(\tau,\hat{y})=0,
    \end{cases}
    \label{eq:skill_update}
\end{equation}
here, $\operatorname{Sanitize}$ removes non-generalizable, duplicate, and instance-specific content, keeping the SOP compact.

\definecolor{oursblue}{rgb}{0.94, 0.92, 0.98}

\definecolor{rowgray}{RGB}{245, 245, 245}

\begin{table*}[t]
\centering

\small
\renewcommand{\arraystretch}{0.95}
\setlength{\tabcolsep}{3pt}

\begin{tabular}{llccccc}
\toprule
\textbf{Backbone} &
\textbf{Method} &
\textbf{\textit{Tool-A-O}$\uparrow$} &
\textbf{\textit{Tool-I-O}$\uparrow$} &
\textbf{\textit{Tool-E-M}$\uparrow$} &
\textbf{\textit{Parameters}$\uparrow$} &
\textbf{\textit{Accuracy}$\uparrow$} \\
\midrule
\multirow{4}{*}{GPT-5}
& Earth-Agent~\cite{feng2025earth}
& 71.16 & 60.68 & 45.79 & \textbf{25.91} & 63.16 \\

& GeoEvolver~\cite{dai2026experience}
& 67.52 & 53.11 & 40.14 & -- & 70.85 \\

& OpenEarth-Agent~\cite{dai2026experience}
& -- & -- & -- & -- & 67.61 \\

& \textbf{GeoForge (Ours)}
& \textbf{79.39}
& \textbf{61.98}
& \textbf{47.39}
& 22.20
& \textbf{74.33} \\

\midrule


\multirow{4}{*}{Gemini-2.5-Flash}

& Earth-Agent~\cite{feng2025earth}
& 61.80 & 50.78 & 40.92 & 23.43 & 55.06 \\

& GeoEvolver~\cite{dai2026experience}
& 61.43 & 53.60 & 38.63 & -- & 59.91 \\

& OpenEarth-Agent~\cite{dai2026experience}
& -- & -- & -- & -- & 57.89 \\

& \textbf{GeoForge (Ours)}
& \textbf{80.35}
& \textbf{65.13}
& \textbf{50.95}
& \textbf{25.46}
& \textbf{67.91} \\

\midrule


\multirow{4}{*}{DeepSeek-V3.1}

& Earth-Agent~\cite{feng2025earth}
& 77.98 & 64.33 & 50.01 & \textbf{31.36} & 52.23 \\

& GeoEvolver~\cite{dai2026experience}
& 61.85 & 47.92 & 37.39 & -- & 59.68 \\


& OpenEarth-Agent~\cite{dai2026experience}
& -- & -- & -- & -- & 56.68 \\

& \textbf{GeoForge (Ours)}
& \textbf{83.58}
& \textbf{68.30}
& \textbf{50.99}
& 26.88
& \textbf{77.09} \\

\midrule


\multirow{3}{*}{GPT-4o}

& Earth-Agent~\cite{feng2025earth}
& 66.88 & \textbf{53.20} & 47.47 & 27.92 & 44.94 \\

& GeoEvolver~\cite{dai2026experience}
& 62.71 & 48.93 & 37.45 & -- & \textbf{65.59} \\


& \textbf{GeoForge (Ours)}
& \textbf{78.91}
& 37.31
& \textbf{52.46}
& \textbf{33.81}
& 57.98 \\

\midrule


\multirow{3}{*}{Qwen3-Max}

& Earth-Agent~\cite{feng2025earth}
& 70.14 & 56.02 & 42.74 & \textbf{26.27} & 47.37 \\

& OpenEarth-Agent~\cite{dai2026experience}
& -- & -- & -- & -- & 52.63 \\

& \textbf{GeoForge (Ours)}
& \textbf{79.26}
& \textbf{61.97}
& \textbf{51.28}
& 25.67
& \textbf{69.72} \\












\bottomrule
\end{tabular}
\caption{Performance comparison of different Earth observation agents on Earth-Bench using diverse LLM backbones.  
}
\label{tab:earth-agent}

\end{table*}

\subsection{Evidence-Grounded Execution}

At inference time, GeoForge assembles a compact execution context by concatenating retrieved graph context, action-level experiences, the adapted SOP, and deterministic EO task guidance:
\begin{equation}
    \mathcal{C}_q=\mathrm{Trunc}_{L_C}
    [\mathcal{G}_q\Vert\mathcal{E}_q\Vert\mathcal{S}_q\Vert\mathcal{H}_q].
    \label{eq:context}
\end{equation}
here $\mathcal{H}_q$ is deterministic EO guidance: inspect available data assets first, organize inputs by modality and acquisition time, prefer compatible batch operations, propagate returned product paths, and verify spatial support, temporal alignment, and parameter semantics before aggregation. Scientific comparisons or predictions are delayed until the required data-derived measurements are available.

The agent then executes a standard ReAct transition over the filtered MCP tool space:
\begin{equation}
    a_t\sim\pi_{\Theta}(\cdot\mid q,\mathcal{C}_q,h_t,\mathcal{T}_q),
    \qquad
    h_{t+1}=h_t\cup\{a_t,o_t\}.
    \label{eq:react_loop}
\end{equation}

Memory is treated only as a processing prior, and every data-grounded EO question requires a tool call, typically beginning with data inventory. For incomplete responses, GeoForge reuses existing geospatial products and observations and invokes only missing operations. A deterministic-plus-LLM normalizer then maps the response to a valid option:
\begin{equation}
    \hat{y}=
    \begin{cases}
    f_{\rm regex}(r), & f_{\rm regex}(r)\neq\emptyset,\\
    f_{\Theta}(q,\mathcal{Y},r,h_{\rm tail}), & \mathrm{otherwise}.
    \end{cases}
    \label{eq:answer_norm}
\end{equation}
where the normalizer creates no evidence; it only converts responses into the benchmark format.

\subsection{Safety-Gated Self-Evolution}

After execution, GeoForge compacts tool calls, arguments, observations, product paths, and the answer into $\bar{\tau}$. A distiller produces
\begin{equation}
    D_{\Theta}(\bar{\tau},q,\hat{y})=
    (\Delta\mathcal{E},\Delta\mathcal{S},\Delta\mathcal{W}),
    \label{eq:distill}
\end{equation}
where $\Delta\mathcal{E}$ contains at most three lessons, $\Delta\mathcal{S}$ a revised SOP, and $\Delta\mathcal{W}$ at most two workflow candidates. Distillation excludes sample-specific paths, answer shortcuts, external APIs, code, and workflows unsupported by actual tool calls.
The update is safety-gated:
\begin{equation}
\begin{aligned}
    \psi(\tau,\hat{y})=&
    \mathbf{1}[\mathrm{Tools}(\tau)>0]\,
    \mathbf{1}[\phi(\hat{y})=0]\\
    &\mathbf{1}[|\mathrm{Calls}(\tau)|\le L_{\max}]\,
    \mathbf{1}[\max_T n_T(\tau)\le R_{\max}]\\
    &\mathbf{1}[\Delta\mathcal{S}\cap\mathcal{F}=\emptyset],
\end{aligned}
\label{eq:update_gate}
\end{equation}
where $\phi(\cdot)$ is the incomplete-answer detector and $\mathcal{F}$ contains forbidden code-like or non-MCP patterns. The memory transition is
\begin{equation}
    \mathcal{M}_{t+1}=
    \left(
    U_G(\mathcal{G}_t,\Delta\mathcal{W}),
    U_E(\mathcal{E}_t,\Delta\mathcal{E}),
    U_S(\mathcal{S}_t,\Delta\mathcal{S};\psi)
    \right).
    \label{eq:self_evolution}
\end{equation}
where $U_G$ normalizes candidates, compresses tool sequences, and merges nodes using weighted tool-sequence and lexical Jaccard similarity before updating edges and pruning by reliability. $U_E$ inserts deduplicated experiences, while $U_S$ replaces the SOP only when $\psi=1$. 
Appendix provides the full algorithm.

\definecolor{darkblue}{RGB}{0, 0, 139}
\definecolor{rowgray}{RGB}{240, 240, 240}
\definecolor{sectionline}{RGB}{100, 100, 100}

\begin{table*}[!t]
    \centering
    \small
    \renewcommand\arraystretch{0.95}
    \setlength{\tabcolsep}{3pt}
    
    
    \begin{tabular}{@{}l c c c c c@{}}
        \toprule
        \multicolumn{6}{@{}c}{\textbf{ThinkGeo}} \\
        \midrule
        \textbf{Method} &
        \textbf{\textit{Inst. Align.}$\uparrow$} &
        \textbf{\textit{Tool Acc.}$\uparrow$} &
        \textbf{\textit{Arg Acc.}$\uparrow$} &
        \textbf{\textit{Ans.}$\uparrow$} &
        \textbf{\textit{Ans. w/ ImgGen}$\uparrow$} \\
        \midrule
        GPT-4o                   & 82.33 & 67.73 & 34.75 & 9.78  & 20.40 \\
        GPT-4-1106               & 86.49 & 74.05 & \textbf{36.96} & 5.16  & 14.69 \\
        Claude-3.7-Sonnet        & 22.31 & 27.31 & 9.00  & 8.97  & 7.51  \\
        Qwen2.5-7b-Instruct      & 64.88 & 51.04 & 20.08 & 7.34  & 6.40   \\
        GeoEvolver~\cite{dai2026experience}    & 80.12 & 76.58 & 35.42 & 46.88 & 53.74 \\
        \textbf{GeoForge (Ours)} & \textbf{97.27} & \textbf{80.31} & 33.96 & \textbf{60.98} & \textbf{62.43} \\
        \midrule
        \multicolumn{6}{@{}c}{\textbf{GeoPlan-Bench}} \\
        \midrule
        \textbf{Method} &
        \textbf{\textit{Recall$_{key}$}$\uparrow$} &
        \textbf{\textit{Precision$_{key}$}$\uparrow$} &
        \textbf{\textit{F1$_{key}$}$\uparrow$} &
        \textbf{\textit{Structural}$\uparrow$} &
        \textbf{\textit{Holistic}$\uparrow$} \\
        \midrule
        CoT~\cite{wei2022chain}            & 0.40 & 0.51 & 0.42 & 0.55 & 980.87 \\
        ReAct~\cite{yao2022react}          & 0.33 & 0.50 & 0.37 & 0.47 & 962.57 \\
        Plan\&Execute~\cite{wang2023plan}  & 0.38 & 0.55 & 0.43 & 0.47 & 973.52 \\
        Debate~\cite{3692070.3692537}      & 0.62 & 0.57 & 0.57 & 0.62 & 1030.59 \\
        AFlow~\cite{zhang2025aflow}        & 0.39 & 0.62 & 0.44 & 0.68 & 992.60 \\
        EarthAgent-MAS~\cite{li2025designing} & 0.66 & 0.65 & 0.63 & 0.68 & 1068.27 \\
        GeoEvolver~\cite{dai2026experience} & 0.64 & 0.68 & 0.63 & 0.45 & 1057.40 \\
        \textbf{GeoForge (Ours)}     & \textbf{1.00} & \textbf{0.70} & \textbf{0.77} & \textbf{0.79} & \textbf{1100.65} \\
        \bottomrule
    \end{tabular}
    \caption{Performance comparison on ThinkGeo and GeoPlan-Bench. }
    \label{tab:thinkgeo_geoplan}
\end{table*}

\section{Experiments}

\subsection{Experimental Setup}
\paragraph{Datasets.}
We evaluate GeoForge on three agentic geospatial benchmarks: Earth-Bench \citep{feng2025earth}, ThinkGeo \citep{shabbir2025thinkgeo}, and GeoPlan-bench \citep{li2025designing}. Together, they cover executable EO analysis, optical/SAR tool reasoning, and high-level geospatial planning, answer correctness, tool selection, parameter use, visual grounding, and workflow completeness. Details are provided in Appendix.

\paragraph{Evaluation Protocols.}
Following \citet{dai2026experience}, we adopt the official evaluation protocols of each benchmark and report final accuracy alongside trajectory-level metrics including Tool-Any-Order (tool coverage), Tool-In-Order (execution order fidelity), Tool-Exact-Match (full sequence precision), Instruction Alignment, Tool Accuracy, Argument Accuracy, Answer Accuracy, and Answer Accuracy w/ ImgGen (parameter correctness for image generation). Additional fine-grained metrics including Parameter Accuracy, Key Tool Recall/Precision, Path Similarity, and Completeness Score are detailed in the Appendix.

\paragraph{Implementation Details.}
We implement GeoForge using LangChain and LangGraph. The LLM is queried at temperature 0.1 with a 120-second timeout, and tools are served via MCP. We adopt a recursion limit of 80 with at most two continuation attempts. The memory system uses top-3 experience retrieval and top-1 workflow retrieval (threshold 0.6).
Additional details are provided in the Appendix.

\begin{table}[t]
\centering

\footnotesize
\renewcommand{\arraystretch}{0.95}
\setlength{\tabcolsep}{3pt}

\resizebox{0.95\columnwidth}{!}{%
\begin{tabular}{@{}l c c c c c c@{}}
\toprule
\textbf{Graph} & \textbf{Skill} & \textbf{Experience} & \textbf{\textit{Accuracy}} & \textbf{\textit{Tool-A-O}} & \textbf{\textit{Tool-I-O}} & \textbf{\textit{Tool-E-M}} \\
\midrule

& & & 52.23 & 78.66 & 64.50 & 49.58 \\

& \checkmark & \checkmark & 67.02 & 75.69 & 61.74 & 45.09 \\

\checkmark & & \checkmark & 52.66 & 80.96 & 67.39 & 48.57 \\

\checkmark & \checkmark & & 70.21 & 76.45 & 62.58 & 47.07 \\

\checkmark & \checkmark & \checkmark & \textbf{74.33} & \textbf{85.89} & \textbf{70.01} & \textbf{51.18} \\

\bottomrule
\end{tabular}%
}
\caption{Ablation study of GeoForge on Earth-Bench.}
\label{tab:ablation}

\end{table}

\subsection{Overall Performance Comparison}
Table~\ref{tab:earth-agent} compares GeoForge with existing EO agents across five LLM backbones on Earth-Bench. GeoForge achieves the best accuracy on four backbones: GPT-5 at 74.33\%, DeepSeek-V3.1 at 77.09\%, Gemini-2.5-Flash at 67.91\%, and Qwen3-Max at 69.72\%, consistently outperforming Earth-Agent, OpenEarth-Agent, and GeoEvolver. It also improves trajectory-level metrics such as Tool-Any-Order, Tool-In-Order, and Tool-Exact-Match on most backbones, demonstrating more reliable tool selection and execution planning.
These results validate that GeoForge enhances both task accuracy and reasoning quality for Earth observation agents.

Table~\ref{tab:thinkgeo_geoplan} further evaluates GeoForge on ThinkGeo and GeoPlan-Bench. On ThinkGeo, GeoForge achieves the highest Instruction Alignment at 97.27\%, Tool Accuracy at 80.31\%, Answer Accuracy at 60.98\%, and grounded Answer Accuracy at 62.43\%. On GeoPlan-Bench, it obtains the best Recall$_{key}$ of 1.00, F1$_{key}$ of 0.77, Structural score of 0.79, and Holistic score of 1100.65, surpassing both traditional planners and recent EO agents. These results demonstrate that GeoForge improves not only low-level tool execution but also high-level workflow planning for complex EO tasks.

\begin{table}[t]
\centering
\scriptsize
\renewcommand{\arraystretch}{0.95}
\setlength{\tabcolsep}{3pt}

\resizebox{\columnwidth}{!}{%
\begin{tabular}{lcccc}
\toprule
\textbf{Method} & \textit{Spectrum} & \textit{Products} & \textit{RGB} & \textit{Avg.} \\
\midrule

GPT-Agent \cite{achiam2023gpt}      & 45.00 & 31.60 & 45.26 & 40.42 \\

MGX \cite{hong2024metagpt}          & 40.00 & 15.80 & 0.00  & 18.60 \\

Manus \cite{shen2025mind}           & 15.00 & 15.80 & 47.62 & 26.14 \\

Coze \cite{kemppainen2025exploring} & 35.00 & 10.50 & 0.00  & 15.30 \\

Earth-Agent \cite{feng2025earth}    & 50.00 & 42.11 & \textbf{51.43} & 47.84 \\

OpenEarth-Agent~\cite{dai2026experience}    & 51.00 & 45.98 & 1.67 & 32.88 \\

\midrule

\textbf{GeoForge (Ours)} & \textbf{77.00} & \textbf{71.26} & 37.29 & \textbf{61.85} \\

\bottomrule
\end{tabular}%
}
\caption{Comparison with general agents across Spectrum, Products, and RGB modalities. }
\label{tab:general_agents_lite}
\end{table}


\subsection{Ablation Study}
\paragraph{Component Contribution.}
The ablation results in Table~\ref{tab:ablation} show that all three memory components contribute to GeoForge. Without memory, the model achieves only 52.23\% accuracy. Removing Skill causes the largest accuracy drop, from 74.33\% to 52.66\%, highlighting the importance of reusable task-level knowledge. In contrast, removing the Workflow Graph leads to the largest degradation in trajectory quality, reducing Tool-Any-Order, Tool-In-Order, and Tool-Exact-Match from 85.89\%, 70.01\%, and 51.18\% to 75.69\%, 61.74\%, and 45.09\%, respectively. Using all three components yields the best performance, demonstrating that graph-, skill-, and experience-level memories provide complementary guidance for multi-step geospatial reasoning.

\begin{figure}[t]
\centering
\includegraphics[width=\columnwidth]{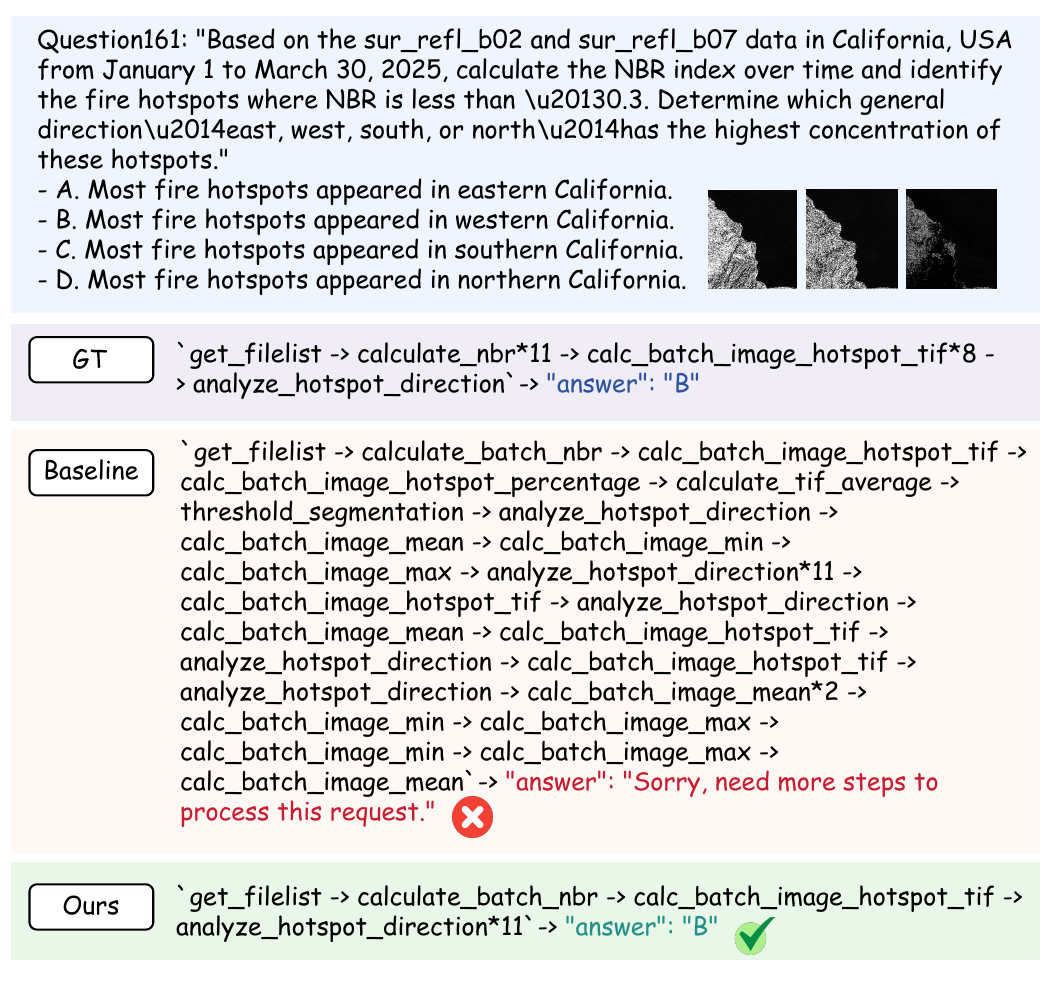} 
\caption{Comparison of tool-use trajectories between the baseline and GeoForge on Earth-Bench. 
}
\label{fig:case_study}
\end{figure}

\begin{figure}[t]
\centering
\includegraphics[width=\columnwidth]{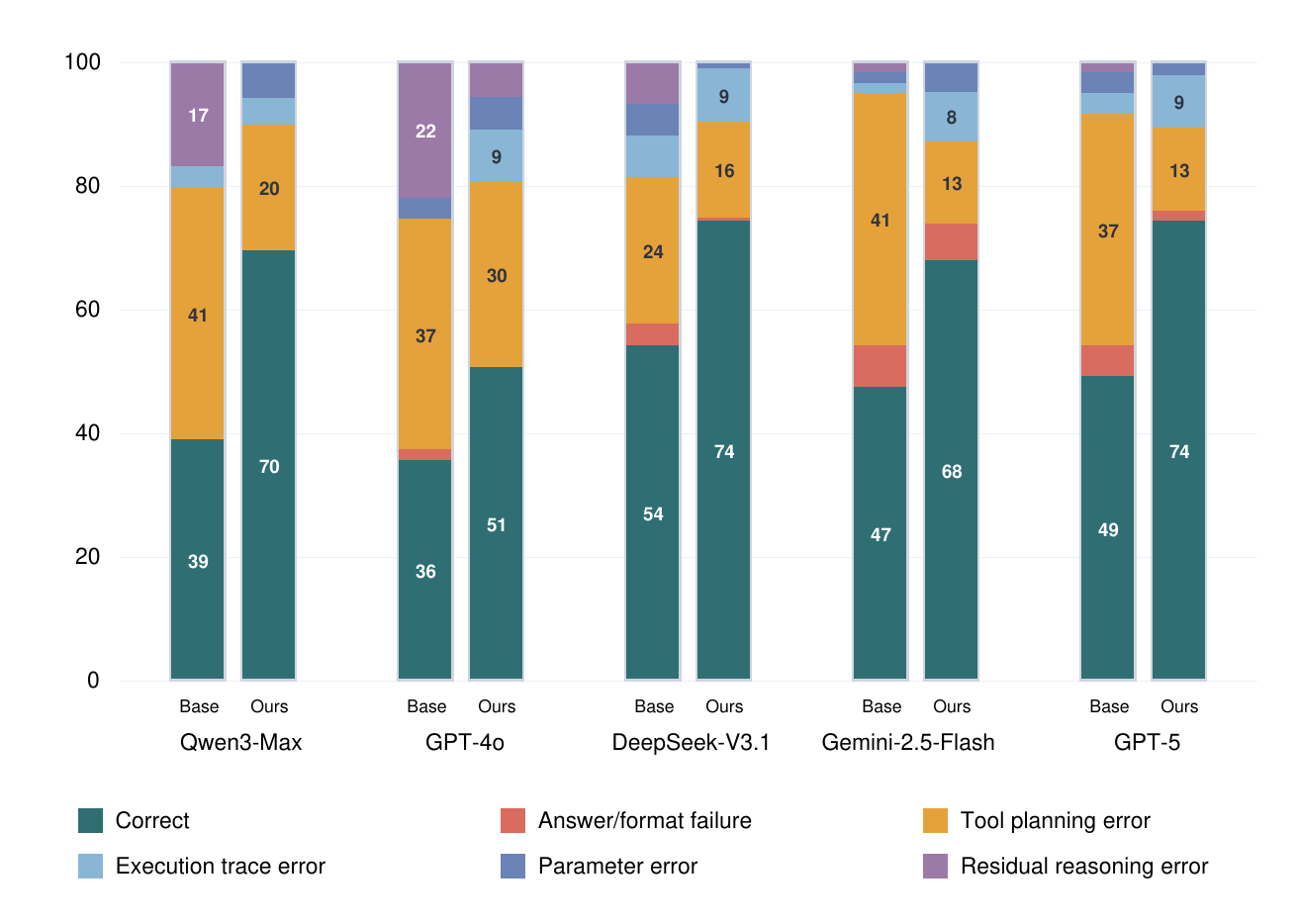} 
\caption{
The percentage distribution of correct predictions and various error types across different backbone LLMs. 
}
\label{fig:error_analysis}
\end{figure}

\begin{figure}[t]
\centering
\includegraphics[width=\columnwidth]{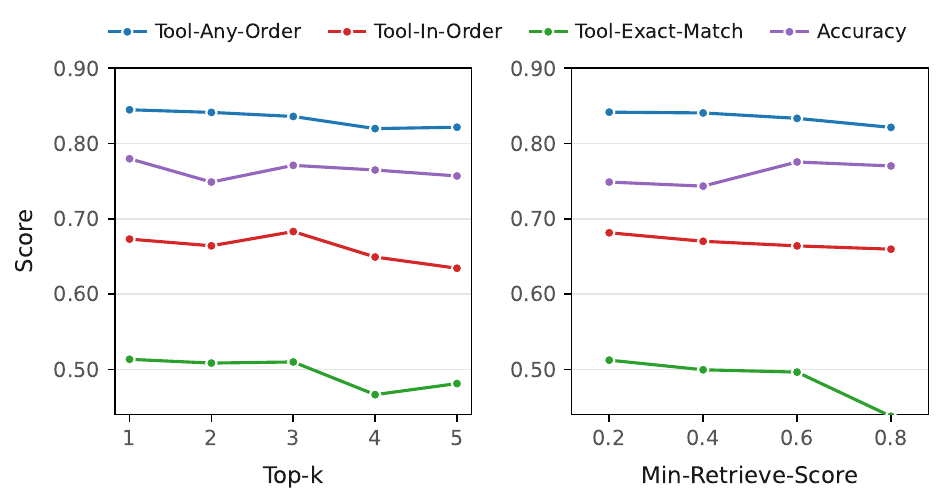} 
\caption{
Sensitivity analysis of the hyperparameters. Left: the effect of the number of retrieved skills (Top-$k$). Right: the effect of the minimum retrieval score threshold. 
}
\label{fig:topk_sensitivity}
\end{figure}

\paragraph{Generalization across Modalities.}
As shown in Table~\ref{tab:general_agents_lite}, GeoForge consistently outperforms all general-purpose agents on Earth-Bench, raising average accuracy from 47.84\% to 61.85\% over the strongest baseline Earth-Agent. It achieves the highest scores on Spectrum at 77.00\% and Products at 71.26\%, with gains of 27.00 and 29.15 percentage points respectively. These substantial improvements on tasks demanding complex spectral processing and multi-step workflows demonstrate that GeoForge's structured memory excels at intricate tool orchestration, validating the effectiveness of its graph-skill-experience architecture for multimodal EO reasoning.

\paragraph{Sensitivity Analysis.} As shown in Figure~\ref{fig:topk_sensitivity}, GeoForge is robust to both the number of retrieved skills and the retrieval score threshold. Retrieving a small number of high-quality skills achieves the best performance, while excessive retrieval slightly degrades trajectory quality due to redundant guidance. Similarly, an overly strict retrieval threshold reduces Tool-Exact-Match by filtering out useful experiences. Therefore, we adopt Top-$k$ = 3 and Min-Retrieve-Score = 0.6, which provide the best trade-off between retrieval quality and planning performance.

\subsection{Error Analysis and Case Study}
Figure~\ref{fig:error_analysis} presents the error distribution across backbone LLMs. Our method consistently improves the correct-answer ratio for all models. These improvements primarily stem from substantial reductions in tool planning errors across most backbones. Reasoning errors are eliminated for the majority of models, and answer or format failures are also notably reduced. Remaining errors are mainly execution-trace and parameter-related issues, suggesting that our method shifts failures from high-level planning toward deeper execution-level challenges.
Figure~\ref{fig:case_study} compares the execution trajectories of the baseline and our method on a representative Earth observation task. Given the task of computing the NBR index, the baseline follows a tangled trajectory with many redundant tool calls, ultimately failing to produce a valid answer. In contrast, our method executes a concise workflow that retrieves the file list, computes the NBR index, detects hotspot regions, and performs a final directional analysis, producing the correct answer B with substantially fewer steps. This demonstrates that GeoForge effectively eliminates redundant actions and maintains task-relevant execution, leading to more reliable problem-solving in complex EO scenarios.

\section{Conclusion}
In this paper, We presented GeoForge, a training-free, self-evolving framework for Earth observation agents that distills completed trajectories into structured nonparametric memory. GeoForge constrains the action space by sensing context and retrieves task-conditioned priors from workflow graphs, action-level experiences, and an adapted skill SOP. Retrieved priors guide execution while observations ground final answers. A safety-gated distillation process enables continuous memory evolution without backbone LLM updates. Experiments across multiple benchmarks show that GeoForge consistently improves task accuracy and tool-use trajectory quality while reducing planning and reasoning errors, offering a scalable paradigm for reliable scientific agents.

\bibliography{aaai2027}


\end{document}